\documentclass[11pt, a4paper, copyright, frontis]{frontis}
\uselogo{}
\fvlogodividerfalse        %
\fvtitlecentertrue         %

\usepackage{latexsym}
\usepackage{iftex}
\ifXeTeX\else
  \PackageError{FrontisBench}{This document must be compiled with XeLaTeX}{Set the Overleaf compiler to XeLaTeX or use the local Tectonic XeTeX build.}
\fi
\usepackage{fontspec}
\defaultfontfeatures{Ligatures=TeX}
\usepackage{xeCJK}
\setCJKsansfont{FandolHei-Regular.otf}[
  BoldFont=FandolHei-Bold.otf
]
\setCJKmonofont{FandolHei-Regular.otf}[
  BoldFont=FandolHei-Bold.otf,
  Scale=MatchLowercase
]
\usepackage{microtype}
\usepackage{graphicx}
\usepackage{tikz}
\usepackage{fontawesome5}
\usepackage{booktabs}
\usepackage{tabularx}
\usepackage{array}
\usepackage{multirow}
\usepackage{makecell}
\usepackage{adjustbox}
\usepackage{enumitem}
\usepackage{listings}
\usepackage{fvextra}
\usepackage{alltt}
\usepackage{mdframed}
\usepackage{xcolor}
\usepackage{amsmath}
\usepackage{url}
\usepackage[numbers,sort]{natbib}
\usepackage{float}
\usepackage{stfloats}
\usepackage{needspace}

\makeatletter
\@ifundefined{NoHyper}{\newenvironment{NoHyper}{}{}}{}
\makeatother
\newcommand{\ecbincludegraphics}[2][]{%
  \begin{NoHyper}%
  \includegraphics[#1]{#2}%
  \end{NoHyper}%
}


\graphicspath{{assets/case_studies/}}

\newcommand{\ecb}{EnterpriseClawBench}
\newcommand{\ecbpipe}{EnterpriseClawBench pipeline}
\newcommand{\hm}{harness--model}
\newcommand{\liteset}{Lite}
\newcommand{\litedesc}{manually audited 120-task \liteset{} subset}
\newcommand{\fullset}{852-task set}
\newcommand{\inlinett}[1]{{\footnotesize\texttt{#1}}}

\definecolor{badgebg}{HTML}{E5F6F8}
\definecolor{badgeborder}{HTML}{8DCFD6}
\newsavebox{\badgeProj}\newsavebox{\badgeGH}\newsavebox{\badgeHF}
\newcommand{\buildbadge}[2]{%
  \tikz[baseline=(t.base)]{
    \node[fill=badgebg, draw=badgeborder, rounded corners=4pt,
          inner xsep=8pt, inner ysep=4pt] (t)
      {#1\kern0.4em{\fontfamily{ppl}\selectfont \small #2}};}}
\savebox{\badgeProj}{\buildbadge{\faGlobe}{Leaderboard}}
\savebox{\badgeGH}{\buildbadge{\faGithub}{GitHub}}
\savebox{\badgeHF}{\buildbadge{%
  \raisebox{-1pt}{\includegraphics[height=0.95em]{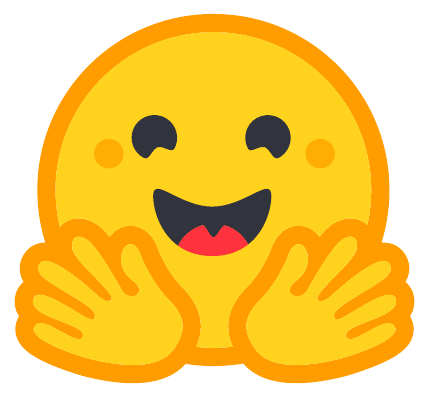}}}{HuggingFace}}

\lstdefinestyle{fbtext}{
  basicstyle=\ttfamily\scriptsize,
  breaklines=true,
  breakatwhitespace=false,
  breakautoindent=false,
  breakindent=0pt,
  columns=fixed,
  basewidth=0.52em,
  keepspaces=true,
  frame=single,
  framerule=0.2pt,
  xleftmargin=1mm,
  xrightmargin=1mm
}
\definecolor{fbcaseblue}{HTML}{5A7FA8}
\definecolor{fbcasefill}{HTML}{F7FAFD}
\definecolor{fbcaseborder}{HTML}{B8CBDD}
\newcommand{\caseheader}[1]{\noindent\colorbox{fbcaseblue}{\parbox{0.96\linewidth}{\color{white}\bfseries #1}}\vspace{0.5mm}\newline}
\newenvironment{casebox}{%
  \begin{quote}%
  \setlength{\parindent}{0pt}%
  \setlength{\parskip}{0pt}%
}{%
  \end{quote}%
}
\newenvironment{caselines}{%
  \begingroup
  \parindent0pt
  \parskip0pt
  \obeylines
  \obeyspaces
}{%
  \endgroup
}
\newcommand{\badgerow}{%
  \makebox[\linewidth][c]{%
    \href{https://frontisai.github.io/EnterpriseClawBench/}{\usebox{\badgeProj}}\hspace{0.8em}%
    \href{https://github.com/FrontisAI/EnterpriseClawBench}{\usebox{\badgeGH}}\hspace{0.8em}%
  }}
\title{EnterpriseClawBench: Benchmarking Agents from Real Workplace Sessions}

\author{
    Jincheng Zhong$^{*}$, Weizhi Wang$^{*}$, Che Jiang$^{*}$ \\ Kai Tian,
    Zhenzhao Yuan, Junlin Yang, Dianqiao Lei, Kaiyan Zhang$^{\ddagger}$
    \vspace{1mm} \\
    \textbf{Horizon Research, Frontis.AI}
    \vspace{1mm} \\
    $^*$ \textbf{Core Contributors.}~~
    \textbf{$^\ddagger$ Corresponding Author.}
    \vspace{1mm} \\
    \texttt{zhangkaiyan@frontis.cn}
    \\ \rule{0pt}{2.0em}\badgerow
  \vspace{-6pt}%
}

\begin{document}

\begin{abstract}
Enterprise agents increasingly operate inside workspaces: they read heterogeneous files, invoke tools, and deliver business artifacts. We introduce \ecb{}, an enterprise agent benchmark constructed from proprietary, real-world agent sessions. Starting from a large archive of workplace sessions, the \ecbpipe{} produces 852 reproducible tasks, each paired with recovered fixtures, rewritten prompts, role classes, skill subclasses, hard rules, and semantic rubrics. Because the sessions contain internal enterprise content, we do not release the benchmark data; instead, our reusable contribution is the construction and evaluation protocol. On \litedesc{}, the best configuration reaches only 0.663 (Codex with GPT-5.5). These results show that enterprise agent evaluation must report harness--model combinations, artifact delivery, visual quality, cost, runtime, and skill-transfer behavior, rather than collapsing performance into a single score.
\end{abstract}
\maketitle

\begin{figure*}[hb]
\centering
\ecbincludegraphics[width=\textwidth]{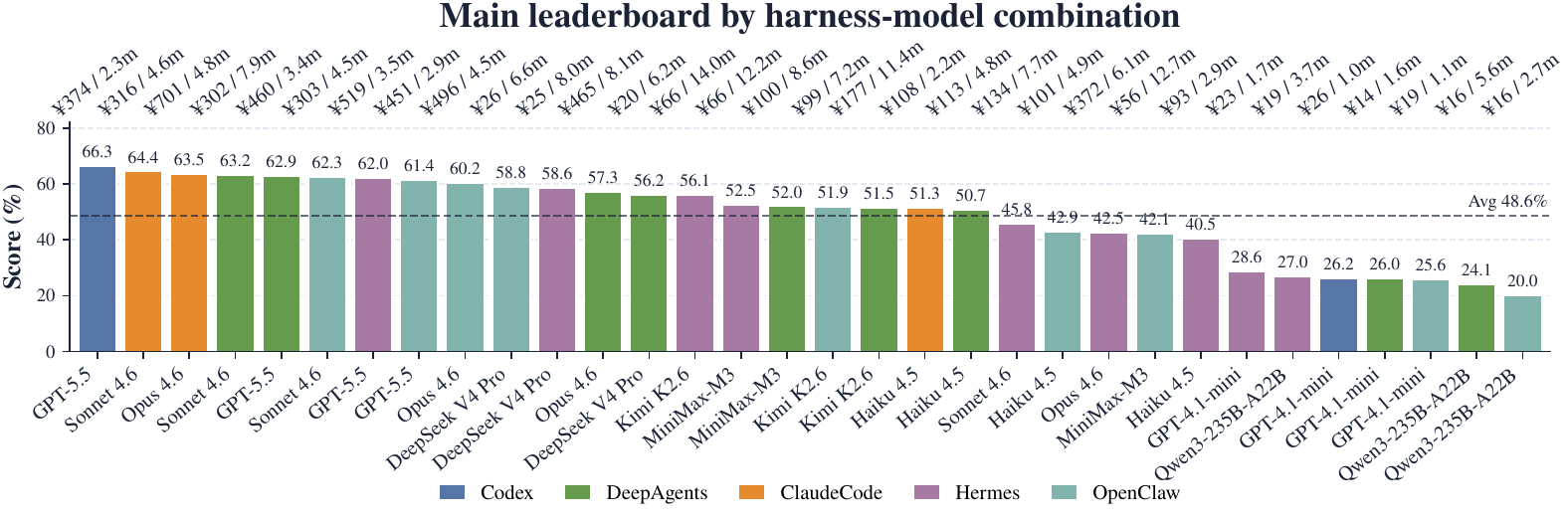}
\caption{Main \liteset{} leaderboard across 32 harness--model combinations. Bar height is score; top annotations show agent cost in CNY and average runtime per task.}
\label{ecblink:fig:leaderboard}
\end{figure*}

\section{Introduction}
Large language models are evolving from text-only assistants into agents that operate inside executable workspaces and return business artifacts. This shift is driven in part by open agent harnesses such as OpenClaw\footnote{https://github.com/openclaw/openclaw}, whose rapid workplace adoption has made the \emph{harness}, the scaffolding that turns a base model into a tool-using agent, a first-class evaluation target; following a recent line of Claw-style benchmarks \citep{wildclawbench2026, clawbench2026, clawswebench2026}, we call such harnesses \emph{claws}. We use \emph{session} for a bounded workplace interaction comprising chat turns, uploaded files, tool traces, generated artifacts, and persistent workspace state \citep{workspacebench2026}. A correct chat response is only part of success for a session: the agent must recover the right inputs, preserve workspace state, and produce usable files under practical cost and latency constraints, requirements that separate enterprise-agent evaluation from knowledge QA and isolated tool use.

\begin{table*}[!t]
\centering
\small
\begin{adjustbox}{max width=\textwidth}
\begin{tabular}{lcccccccc}
\toprule
Benchmark & Tasks & Real-source & Artifact types & Multimodal & Skill eval & Cost/time & Reported combos \\
\midrule
Workspace-Bench & 388/100 & Yes & Many & Yes & No & Partly & 28 \\
EnterpriseBench & 500 & No & N/A & No & No & Partly & 10\\
WildClawBench & 60 & No & Many & Yes & Yes & Yes & 31 \\
ClawBench & 153 & No & N/A & Yes & No & No & 7 \\
Claw-SWE-Bench & 350/80 & Yes & N/A & No & No & Partly & 17 \\
SkillsBench & 87 & No & Many & Yes & Yes & Partly & 18 \\
\textbf{\ecb{} (Ours)} & \textbf{852/120} & \textbf{Yes} & \textbf{Many} & \textbf{Yes} & \textbf{Yes} & \textbf{Yes} & \textbf{32} \\
\bottomrule
\end{tabular}
\end{adjustbox}
\caption{Positioning of \ecb{} against related benchmarks.
``Tasks'' follows each paper's stated counts; ``/'' separates full and Lite versions. ``Real-source'' means the underlying sources or traces are produced in real use, not authored for the benchmark. ``Artifact types'' counts expected output file categories; ``N/A'' means the benchmark does not define success through output-file artifacts. ``Skill eval'' means explicit skill
experiments. ``Cost/time'' means reported efficiency metrics. ``Reported combos'' counts reported harness--model pairs.}
\label{ecblink:tab:related}
\end{table*}
\begin{figure*}[!b]
\centering
\ecbincludegraphics[width=\textwidth]{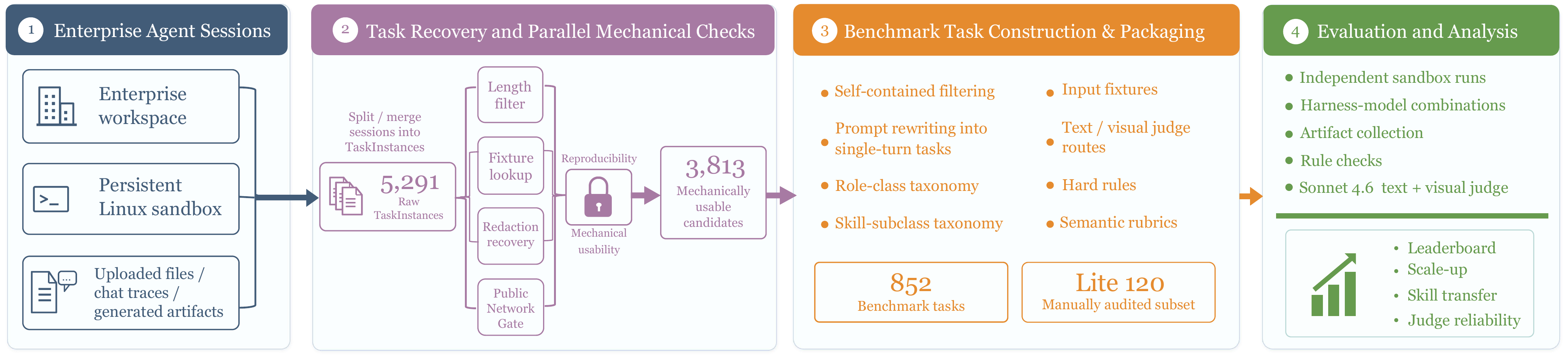}
\caption{Compact overview of the \ecbpipe{} from enterprise sessions to benchmark tasks and evaluation assets.}
\label{ecblink:fig:pipeline}
\end{figure*}
Existing benchmarks have made clear progress toward this setting, but three gaps remain. \textbf{Gap 1: the tension between enterprise realism and scalable task construction.} Workspace-Bench \citep{workspacebench2026}, WorkArena \citep{workarena2024}, TheAgentCompany \citep{theagentcompany2025}, EnterpriseBench \citep{enterprisebench2025}, and EntWorld \citep{entworld2026} move agent evaluation into workspace or enterprise-like environments. However, many tasks are still human-authored, simulated, or built from public environments, leaving a gap between benchmark tasks and naturally occurring enterprise demand. \textbf{Gap 2: multidimensional evaluation.} Recent work has already shown that agent performance is shaped by more than the base model: WildClawBench \citep{wildclawbench2026} reports harness effects, ClawBench \citep{clawbench2026} emphasizes environment- and state-grounded execution, Claw-SWE-Bench \citep{clawswebench2026} adapts Claw-style harness evaluation to repository-level code repair, and AgenticVBench \citep{agenticvbench2026} introduces multimodal production tasks. Enterprise evaluation therefore needs to report \hm{} combinations, multimodal judging, artifact delivery quality, time, and cost as a coupled result rather than a single model score. \textbf{Gap 3: task-class-level skill evaluation.} Reusable skills are becoming operational assets in enterprise agent systems, but existing benchmarks are either not designed for skill evaluation or evaluate skills at the level of individual task items~\citep{skillsbench2026}. Enterprise skills are better tested as transfer units: a skill distilled from one task class should improve held-out tasks from the same class.

We introduce \ecb{}, an internal enterprise benchmark constructed from historical internal agent sessions. The \ecbpipe{} performs reproducibility and privacy checks, fixture recovery, prompt rewriting, taxonomy assignment, deliverable detection, hard-rule generation, semantic rubric generation, and sandbox preflight. Because the source data include proprietary sessions, attachments, and business context, the benchmark data are not released. The paper contributes:
\begin{itemize}[leftmargin=*,itemsep=1pt,topsep=2pt]
  \item an automated construction protocol that converts real enterprise agent sessions into reproducible benchmark tasks;
  \item a multidimensional evaluation framework that jointly reports \hm{} performance, file delivery, text/visual semantic quality, cost, and runtime;
  \item native support for evaluating skill generalization across held-out tasks from the same enterprise task class.
\end{itemize}

\section{\ecb{} Data and Construction}
\paragraph{Enterprise setting.}
\ecb{} is built from continuous internal use of an enterprise agent system at an AI startup with more than 100 employees, covering workplace sessions from March to May 2026. Employees interact with the agent through private or group chats in an enterprise collaboration platform, upload files, and expect deliverables to appear in a persistent Linux workspace. Within each session, inputs are mounted under an input directory, generated files must be written to an output directory, and the platform converts these outputs into downloadable artifacts. This deployment setting provides naturally occurring employee requests, logged fixtures, tool traces, and delivered files, which are the raw materials for constructing an enterprise-oriented benchmark.

\begin{figure}[H]
\centering
\ecbincludegraphics[width=0.5\columnwidth]{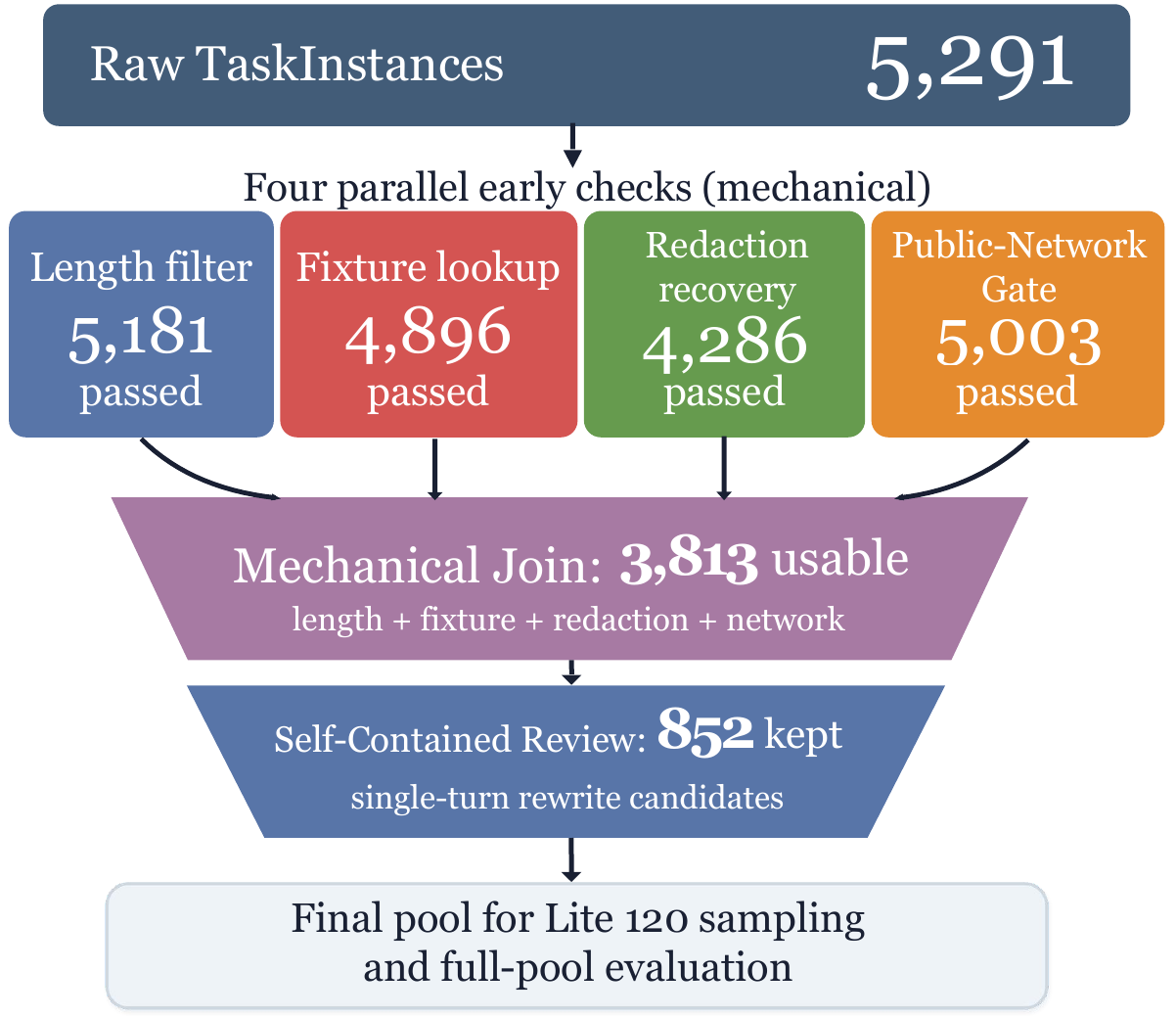}
\caption{Construction funnel from raw TaskInstances to final benchmark tasks.}
\label{ecblink:fig:funnel}
\end{figure}

\paragraph{Construction pipeline.}
The \ecbpipe{} converts noisy proprietary sessions into reproducible benchmark tasks (Figure~\ref{ecblink:fig:pipeline}). It first recovers task-level candidates by splitting and merging session turns. It then applies mechanical gates, including length filtering, fixture lookup, redaction recovery, and network-dependency checks. Candidates that pass these reproducibility gates are judged for self-containment, rewritten into single-turn benchmark prompts, assigned role classes and skill subclasses, annotated with expected deliverables, and packaged with hard rules and text/visual semantic rubrics. Figure~\ref{ecblink:fig:funnel} gives the main reduction path. A final sandbox preflight checks input upload, agent execution, artifact download, and judge routing before a task enters the benchmark set.

\begin{figure*}[!t]
\centering
\ecbincludegraphics[width=\textwidth]{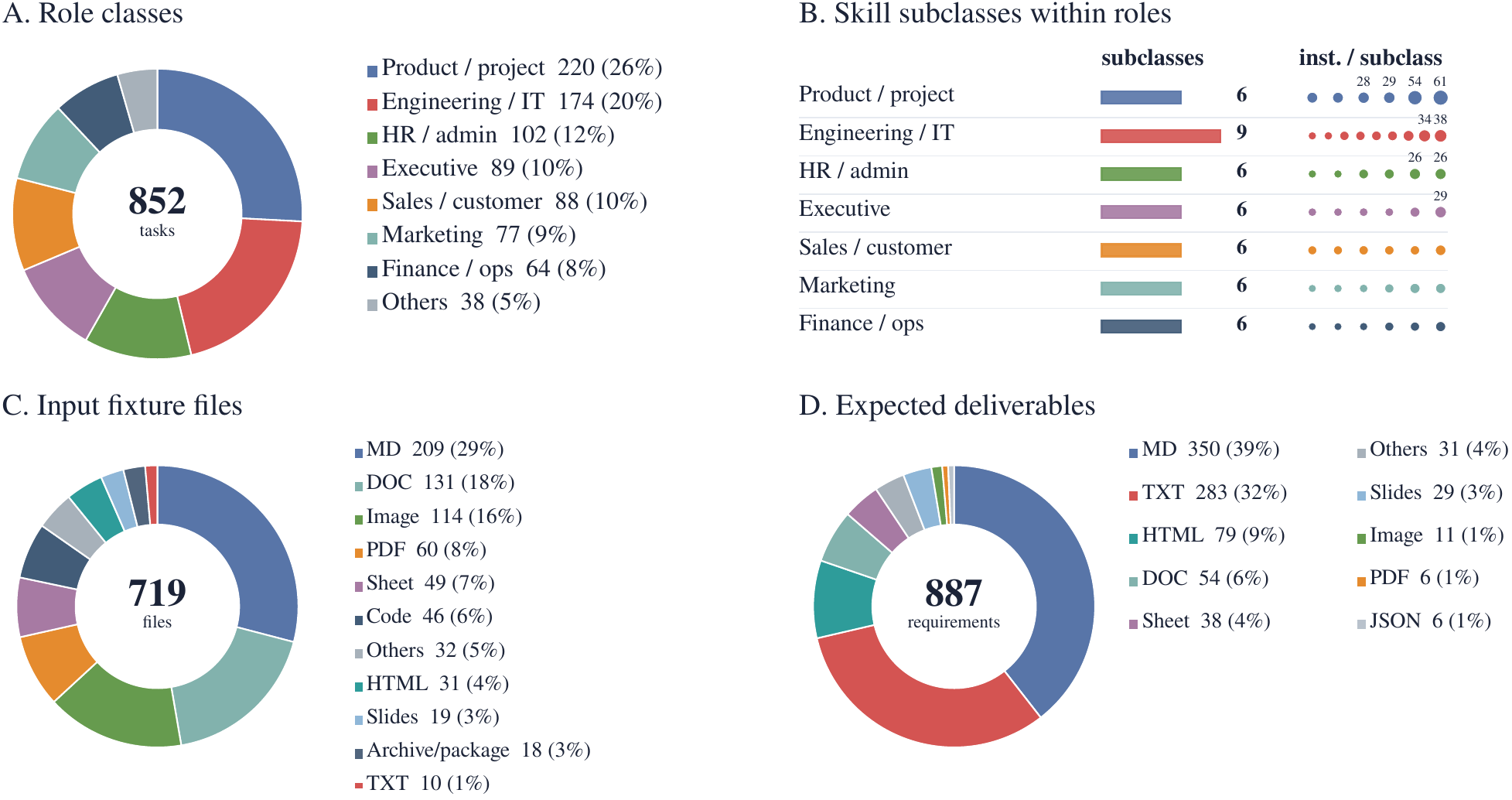}
\caption{Benchmark statistics for the 852-task set. Panel A uses the role-class taxonomy; Panel B shows role-specific skill-subclass structure for the tasks; Panel C and D count input and expected artifact types, respectively.}
\label{ecblink:fig:benchmark-stats-main}
\end{figure*}

\paragraph{Benchmark statistics.}
The construction starts from a large session archive, recovers 5,291 raw TaskInstances, and yields 852 final benchmark tasks. Figure~\ref{ecblink:fig:benchmark-stats-main} shows that the task set is broad but not uniform. Product/project and engineering/IT roles form the largest blocks, while finance, marketing, sales, executive, and HR/admin tasks create a long tail of specialized work. The taxonomy also exposes structure within roles: the seven explicit role classes expand into 45 role-specific skill subclasses, which makes task-class-level skill evaluation possible. Both the input side and the output side are heterogeneous in file type and artifact modality. Note that expected deliverable counts sum to more than 852 because some tasks ask for multiple files or output types.

\section{Experiments}
\subsection{Evaluation Setting}
\ecb{} evaluates \hm{} combinations rather than model names alone. The main leaderboard covers Claude Code, Codex, DeepAgents, Hermes, and OpenClaw with supported models including GPT-5.5, Sonnet 4.6, Opus 4.6, Haiku 4.5, Kimi K2.6, MiniMax-M3, GPT-4.1-mini, Qwen3-235B-A22B, and DeepSeek V4 Pro where available. All tasks run independently in non-stateful sandboxes. The runner uploads task inputs, invokes the harness, downloads outputs and traces, filters temporary files, and records completion, runtime, token use, cost, tool calls, and evidence warnings.

Scoring has two layers. Hard rules check objective delivery properties such as required file type, file count, non-emptiness, openability, tracebacks, and unreplaced placeholders. Semantic judges then score output quality along five dimensions: grounded accuracy, task relevance, substantive depth, practical utility, and communication quality, drawing on multi-metric LLM-evaluation practice \mbox{\citep{helm2022,geval2023,ragas2023}}. The semantic layer is routed by modality: text-extractable outputs go to a text judge, while HTML, slides, PDFs, spreadsheets, and images are rendered into screenshots or page images for a visual judge.

\subsection{Main Leaderboard on \liteset{}}
The primary result uses \litedesc{}. The main 32-combo score uses Sonnet 4.6 as both text and visual judges. Figure~\ref{ecblink:fig:leaderboard} reports the full leaderboard. The best score is only 0.663 for Codex/GPT-5.5, showing that enterprise artifact tasks remain far from saturated.

\paragraph{Harness--model interaction.}
The leaderboard is shaped by model--harness coupling. The clearest pattern is the Claude-family drop under Hermes. Sonnet 4.6 stays in the 0.62--0.64 band under Claude Code, DeepAgents, and OpenClaw, then falls to 0.458 under Hermes. Opus 4.6 and Haiku 4.5 follow the same direction.   Trace inspection suggests a runtime-level mismatch: Claude-family models often rely on active environment probing, script execution, and multi-step repair,while Hermes more frequently blocks these actions through approval checks, routes blocked work into delegated subtasks, or returns long execution traces that are
truncated before the artifact-writing loop is completed. As a result, a session may finish without a stable file under \inlinett{/workspace/outputs}. The same tasks are often completed by Claude Code or DeepAgents with the same Claude model, indicating a \hm{} compatibility issue rather than a model-only capability gap.

\begin{figure*}[t]
\centering
\ecbincludegraphics[width=\textwidth]{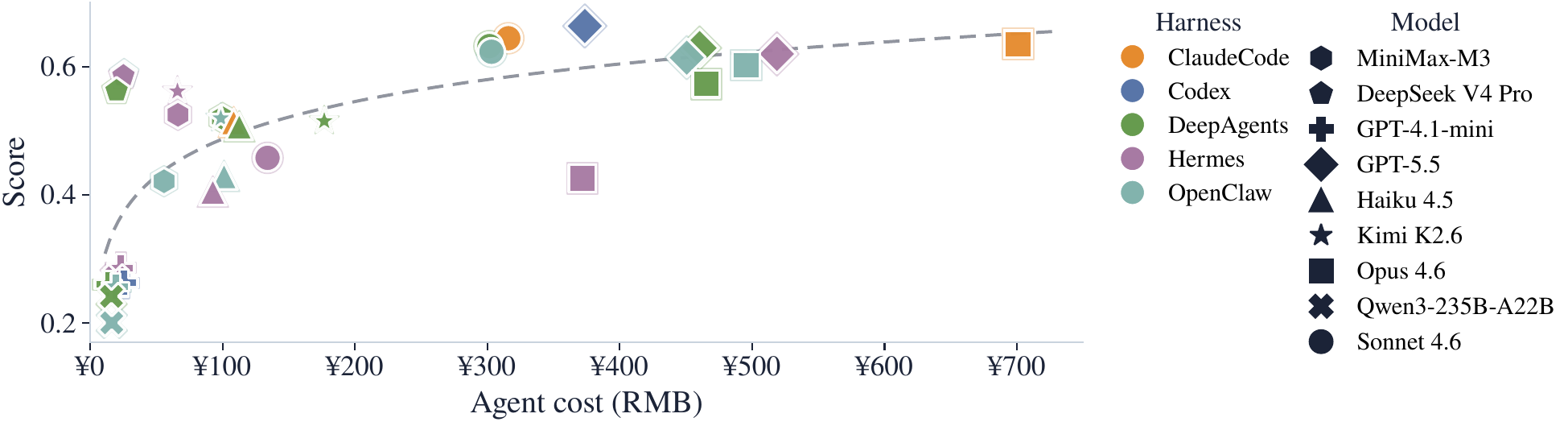}
\caption{Cost--score trade-off on \liteset{}. Each point is one harness--model combination; color encodes the harness and marker shape encodes the model. The score follows the main leaderboard scoring protocol.}
\label{ecblink:fig:score-cost}
\end{figure*}

\paragraph{Cost--score trade-off.}
The full cost--score scatter is provided in Figure~\ref{ecblink:fig:score-cost}. Higher agent cost is generally associated non-linearly with higher score. Moving from very low-cost systems to mid-cost systems brings large gains; beyond the mid range, additional cost produces smaller improvements, giving the scatter a log-like shape. This pattern matches the deployment intuition that stronger models buy reliability and artifact quality, but with diminishing marginal returns. The main outliers are the Hermes/Claude-family combinations, where high model cost does not translate into high score because of the harness--model interaction issue.

\begin{figure}[t]
\centering
\ecbincludegraphics[width=0.65\columnwidth]{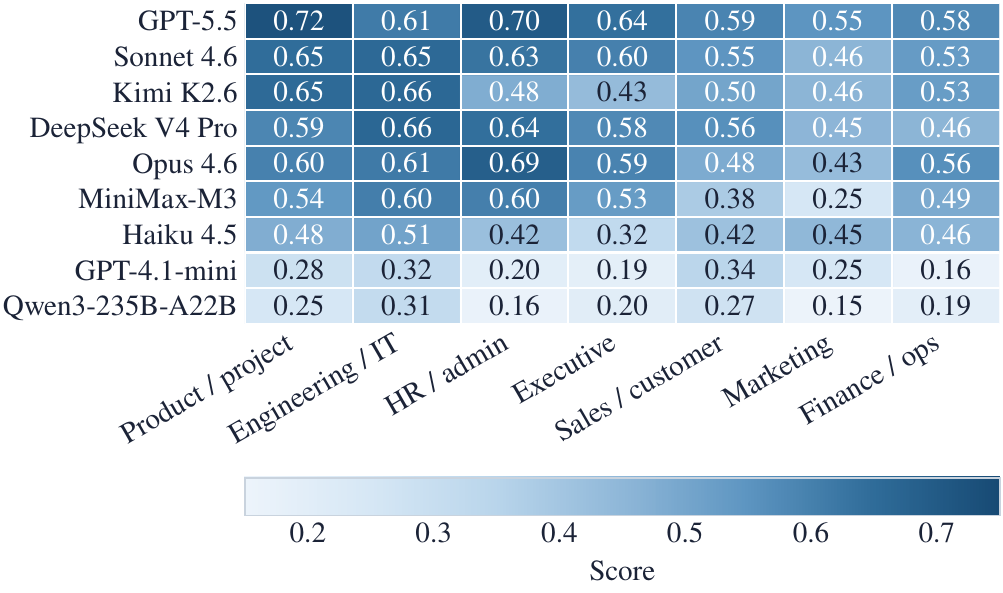}
\caption{Model performance differs by enterprise role class. Columns correspond to organizational functions in which the original tasks occurred.}
\label{ecblink:fig:role-class-main}
\end{figure}

\paragraph{Role-class effects.}
The role-class taxonomy groups tasks by the real enterprise roles, as shown in Figure~\ref{ecblink:fig:role-class-main}. The heatmap shows clear patterns. Among all models, GPT-5.5 is the most robust generalist with leading performance across multiple role classes. Among role classes, marketing and finance/operations remain harder. Manual inspection of the tasks and traces suggests that these tasks combine heavy document comprehension and generation with company-specific scenarios and conventions. These tasks are less common in public data, which may partly explain their lower scores relative to other classes.

\begin{figure}[t]
\centering
\ecbincludegraphics[width=0.65\columnwidth]{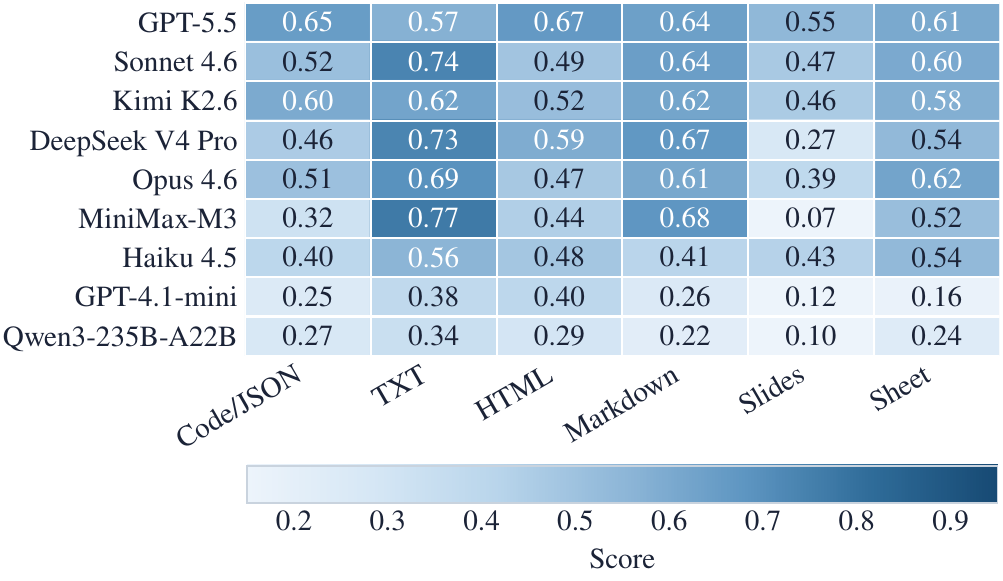}
\caption{Model performance differs by expected deliverable type.}
\label{ecblink:fig:artifact-main}
\end{figure}

\paragraph{Artifact-type effects.}
Expected deliverables created a second evaluation axis. Figure~\ref{ecblink:fig:artifact-main} shows that file format changes the ranking. GPT-5.5 is strongest on HTML, code/JSON, and several delivery-heavy categories, while Opus 4.6 is strongest on spreadsheets. We also observe a systematic score inflation on visually judged spreadsheet and presentation artifacts compared with text-only deliverables. The same pattern reappears in the judge ablation study, suggesting that current multimodal LLM judges remain less calibrated than text judges for fine-grained enterprise artifact evaluation.

\begin{figure}[t]
\centering
\ecbincludegraphics[width=0.69\columnwidth]{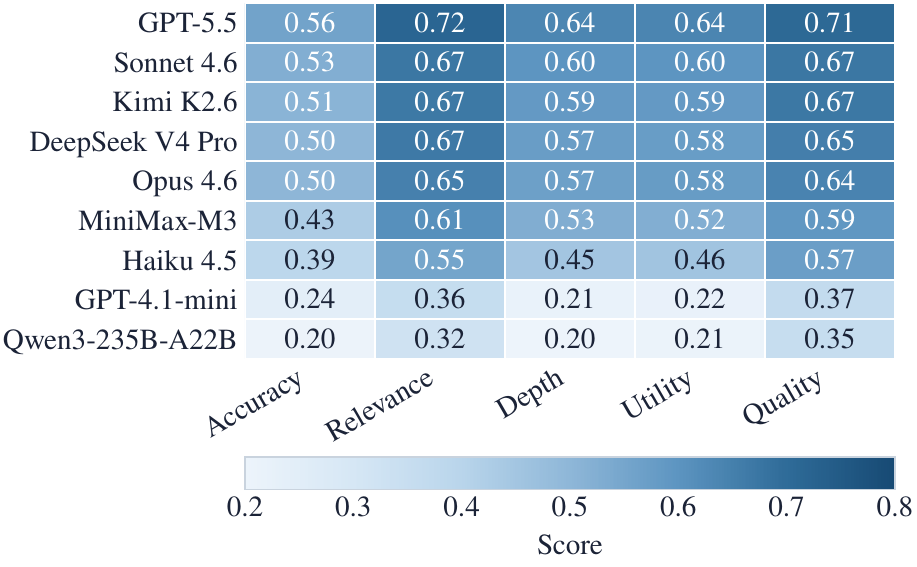}
\caption{Five-dimension semantic diagnostic by model.}
\label{ecblink:fig:radar-main}
\end{figure}

\paragraph{Rubric-dimension analysis.}
The five semantic dimensions identify quality bottlenecks. Current systems were generally better at communication quality and task relevance than at grounded accuracy. This pattern is consistent with enterprise tasks that require reading large input files and carrying evidence into business artifacts. Manual inspection suggests that agents often fail to locate the key information requested by the user in large uploaded files, or lose such information during long-horizon execution. Figure~\ref{ecblink:fig:radar-main} shows the model-level diagnostic profile.

\subsection{Scalability Check on the Full Set}
We run four DeepAgents combinations on the automatically constructed \fullset{} as a scalability check for the \ecbpipe{}. As shown in Table~\ref{ecblink:tab:scalability}, GPT-5.5 remains the strongest model, Sonnet 4.6 forms the second tier at a close distance, Haiku 4.5 remains intermediate, and GPT-4.1-mini is clearly weaker, which aligns with the main leaderboard. The full benchmark produced by the scalable automated pipeline preserves the broad model ranking observed on the manually audited \liteset{} subset, while remaining unsaturated even for the strongest models and retaining clear score separation, supporting the effectiveness of the pipeline itself.
\begin{table}[h]
\centering
\small
\begin{tabular}{lrrrr}
\toprule
Model & Score & Text & Visual & Rule \\
\midrule
GPT-5.5 & 0.766 & 0.813 & 0.642 & 0.959 \\
Sonnet 4.6 & 0.749 & 0.793 & 0.634 & 0.957 \\
Haiku 4.5 & 0.632 & 0.666 & 0.542 & 0.963 \\
GPT-4.1-mini & 0.336 & 0.383 & 0.213 & 0.817 \\
\bottomrule
\end{tabular}
\caption{Scalability check on the full 852-task set under the DeepAgents harness.}
\label{ecblink:tab:scalability}
\end{table}

\subsection{Skill Evaluation}
For skill evaluation, we focus on one task subclass, frontend page generation. We first run each consumer agent on 10 in-domain tasks from this subclass and collect its execution traces, delivered artifacts, and Sonnet 4.6 judge feedback. These materials are then provided to a skill creator, which distills an agent-specific skill for the corresponding consumer. The generated skill is injected back into the same consumer agent, and we evaluate the agent on 5 held-out tasks from the same subclass. The reported effect is the change in the held-out average score before and after skill injection, measuring whether a skill distilled from in-domain executions transfers to new tasks within the same task class.

\begin{figure}[h]
\centering
\ecbincludegraphics[width=0.6\columnwidth]{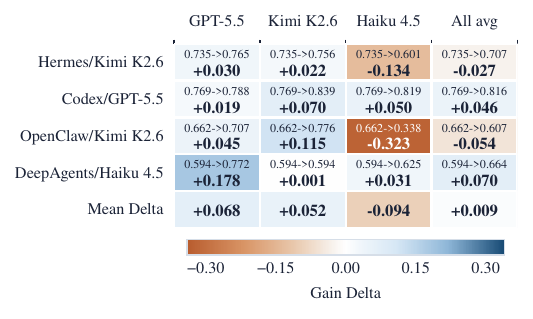}
\caption{Task-class skill injection matrix for held-out frontend-page-generation tasks. Each creator cell reports no-skill score $\rightarrow$ skill-injected score and the corresponding score delta.}
\label{ecblink:fig:skill-transfer-main}
\end{figure}

The matrix shown in~\ref{ecblink:fig:skill-transfer-main} suggests that skill quality depends strongly on the creator model. GPT-5.5 is the strongest creator in this experiment, producing positive average transfer (+0.0681) and no negative delta. Kimi K2.6 is also positive on average (+0.0518), although its gains are concentrated in strong consumers like the Codex/GPT-5.5 and OpenClaw/Kimi K2.6. Haiku 4.5 is the weakest creator: it produces a negative mean delta (-0.0941) and a large degradation for OpenClaw/Kimi K2.6.

Skill-creation ability and skill-consumption ability are not necessarily aligned. Haiku 4.5 is a weak creator, but DeepAgents/Haiku 4.5 is a capable consumer in this task class: two out of three injected skills improve the baseline. Conversely, stronger consumers can still be sensitive to creator choice. Codex/GPT-5.5 gains under all three creators, while Hermes/Kimi K2.6 and OpenClaw/Kimi K2.6 are hurt by Haiku-created skills.

Skill injection is therefore high variance. The outcome depends on creator quality, consumer behavior, creator--consumer fit, and the baseline score of the consumer. These results support evaluating skills at the task-class level with a consumer--creator matrix rather than reporting a single averaged skill score.

\subsection{Judge Reliability}
Judge reliability has two layers: agreement among LLM judges and agreement between the main LLM judge and human raters. For LLM--LLM agreement, we rescore the same \liteset{} artifacts across 32 \hm{} combinations. GPT-5.4-text correlates strongly with the Sonnet 4.6 main text judge ($\rho=0.918$ over 1,853 cases), and GPT-5.4-visual also preserves the broad ordering of visual-route scores ($\rho=0.866$ over 1,428 cases). Appendix~\ref{ecblink:sec:judge-ablation} provides the full ablation analysis.

For human agreement, we audit 48 packets split evenly between text and visual routes (Table~\ref{ecblink:tab:human-judge}). The text route aligns well with human scores: means are close, MAE is 0.134, and Spearman correlation is 0.790. Visual agreement is much weaker: Sonnet 4.6 is substantially more conservative than humans, with MAE 0.303 and negative rank correlation. This is an important gap exposed by \ecb{}: evaluation on multimodal artifacts is not yet mature. 
\begin{table}[ht]
\centering
\small
\setlength{\tabcolsep}{3.6pt}
\begin{tabular}{lrrrrr}
\toprule
Scope & $n$ & Human & Sonnet & MAE & Spearman \\
\midrule
Overall & 48 & 0.571 & 0.498 & 0.219 & 0.263 \\
Text & 24 & 0.476 & 0.504 & 0.134 & 0.790 \\
Visual & 24 & 0.666 & 0.492 & 0.303 & -0.259 \\
\bottomrule
\end{tabular}
\caption{Human calibration audit for the Sonnet 4.6 judge. Human and Sonnet columns report mean scores.}
\label{ecblink:tab:human-judge}
\end{table}

\section{Conclusion}
We presented \ecb{}, an internal enterprise agent benchmark built from real internal agent sessions. The \ecbpipe{} automatically converts proprietary workplace archive into 852 reproducible, artifact-centric tasks with fixtures, taxonomy, and text/visual rubrics. The tasks remain unsaturated, and model performance changes substantially with the harness used. Beyond final scores, cost, runtime, artifact delivery, and multi-dimensional semantic quality are necessary to explain system behavior. The task-class taxonomy also makes it naturally suited for skill-generalization testing.

\section{Limitations}
\ecb{} is built from one enterprise deployment, so the results may not represent all organizations. The full dataset is not released because it contains internal sessions, attachments, tool traces, and business artifacts. Finally, LLM judges remain imperfect, especially for visual artifacts; judge ablation and human-correlation experiments are therefore necessary calibration steps.

\section{Ethical Considerations}
Any organization reusing this approach should obtain proper authorization, restrict data access, review sensitive information, preserve redaction and recovery audit records, and ensure that benchmark construction does not leak employees, customers, or proprietary information.

\bibliographystyle{acl_natbib}
\bibliography{custom}
\appendix

\section{Appendix: Judge Ablation}
\label{ecblink:sec:judge-ablation}
\begin{figure*}[!h]
\centering
\ecbincludegraphics[height=0.92\textheight]{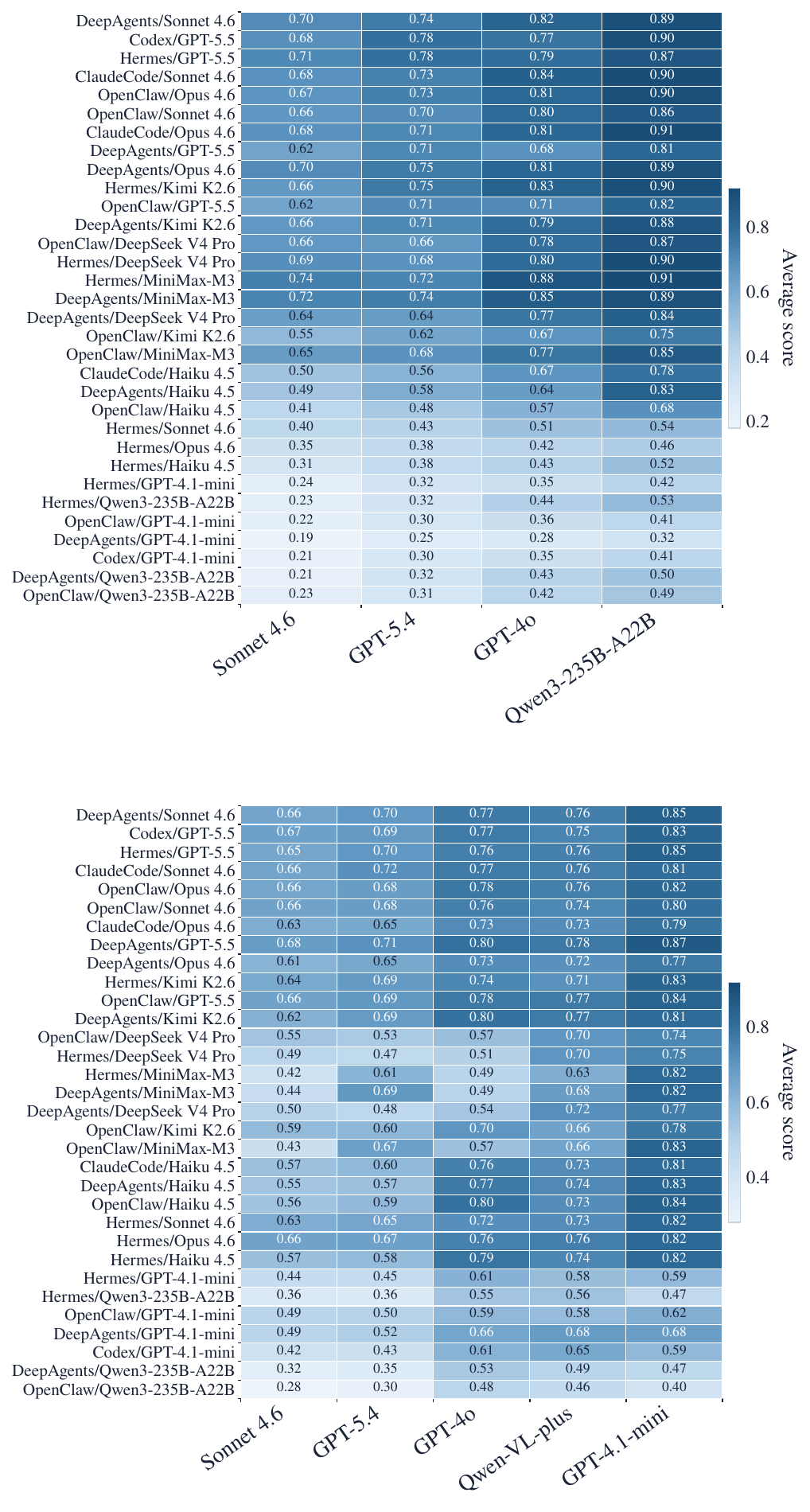}
\caption{Judge-model ablation over \liteset{} artifacts. Top: text route; bottom: visual route; rows sorted by Sonnet 4.6-main score.}
\label{ecblink:fig:judge-appendix}
\end{figure*}

The main text reports the judge-reliability conclusion. Here we separate the LLM-judge analysis into horizontal and vertical views. Horizontally, different judges use different absolute score scales; Strong models are generally more conservative, while less-capable judges assign higher scores. Scores from different judges should therefore not be mixed in one leaderboard. Vertically, capable judges preserve the broad ordering of the main leaderboard: strong systems remain in the upper tier and weak systems remain in the lower tier. The largest instability appears in the middle of the ranking and is more visible for visual artifacts than for text artifacts. Figure~\ref{ecblink:fig:judge-appendix} gives the full 32-combo ablation heatmaps for audit, split into text and visual judge routes.

\section{Appendix: Case Studies}
This appendix provides release-safe case studies. Internal entities, URLs, employee identifiers, customer identifiers, project names, session identifiers, and monetary values are masked as \inlinett{****}. Each evidence box preserves the original Chinese excerpt and adds a literal English translation inside the same box. These cases are included to show the actual task packets, model outputs, skill files, and judge feedbacks behind the benchmark claims.

\subsection{End-to-End Construction Case}
This section shows intermediate examples from the \ecbpipe{} as it turns raw TaskInstances into usable benchmark instances.

\caseheader{Length filtering.}
The raw TaskInstance \inlinett{fb-****\_i****} contains only a single-character user message:

\begin{casebox}\scriptsize\raggedright
\begin{caselines}
Original Chinese:\nopagebreak[4]
benchmark\_id: fb-****\_i****
business\_task: 信息不足无法分类
user\_messages:
- ？
cleaning.length:
  status: rejected
  effective\_user\_chars: 1
  min\_user\_chars: 10

---
English translation:\nopagebreak[4]
benchmark\_id: fb-****\_i****
business\_task: system-error handling
user\_messages:
- ?
cleaning.length:
  status: rejected
  effective\_user\_chars: 1
  min\_user\_chars: 10
\end{caselines}
\end{casebox}

Because the sample contains too little linguistic information to recover a task objective, it is removed by the earliest length filter. This low-cost rule eliminates empty messages, extremely short messages, and system noise before more expensive recovery stages.

\caseheader{Input-fixture recovery.}
The raw TaskInstance \inlinett{fb-****\_i****} declares a DOCX transcript attachment, but the pipeline cannot recover a reproducible file from the candidate source paths:

\begin{casebox}\scriptsize\raggedright
\begin{caselines}
Original Chinese:\nopagebreak[4]
benchmark\_id: fb-****\_i****
business\_task: 生成****评估报告
user\_messages:
- 员工****: 用****自动评估助手分析这个录音

  附件: /root/inputs/****\_****录音.docx
- 员工****: 从聊天录音上来看，并结合现在市场上的****情况，你建议我们这边给到****
- 员工****: 重新搜索录音记录，他一开始期望的是****，后面因为公司是***，他觉得可以按照****来
- 员工****: 而且他也没有外部****，重新生成一下****评估

declared fixture:
  source\_path: /root/inputs/****\_****录音.docx
  target\_path: inputs/****\_****录音.docx
  type: file

cleaning.fixture:
  status: rejected
  declared: 1
  resolved: 0
  total\_bytes: 0

---
English translation:\nopagebreak[4]
benchmark\_id: fb-****\_i****
business\_task: generate a evaluation report on ****
user\_messages:
- Employee ****: use the **** automatic evaluation assistant to analyze this recording.
  Attachment: /root/inputs/****\_**** recording.docx
- Employee ****: based on the recording and current market conditions, what would you suggest we offer?
- Employee ****: search the recording again. The initial expectation was ****, but later, because the company is ***, the task should follow ****.
- Employee ****: the person has no external **** either. Regenerate the **** evaluation.

declared fixture:
  source\_path: /root/inputs/****\_**** recording.docx
  target\_path: inputs/****\_**** recording.docx
  type: file

cleaning.fixture:
  status: rejected
  declared: 1
  resolved: 0
  total\_bytes: 0
\end{caselines}
\end{casebox}

The user prompt is understandable, but the required input file is missing. Keeping this instance would mix data-availability failure with agent ability, so the fixture gate removes it.

\caseheader{Redaction recovery.}
Enterprise traces often contain redacted URLs or file paths. In this example, a redacted link in the user message can be restored with high confidence from the same session context:

\begin{casebox}\scriptsize\raggedright
\begin{caselines}
Original Chinese:\nopagebreak[4]
instance\_id: ****\_i****
field\_path: user\_messages[0]
redacted\_value: https://****/[redacted]
restored\_value: https://****/record?id=****
status: restored\_by\_context
confidence: high
match\_strategy: fuzzy\_context
source\_preview: [13:02:23] 使用****自动评估技能，识别这个链接的内容 https://****/record?id=****

---
English translation:\nopagebreak[4]
instance\_id: ****\_i****
field\_path: user\_messages[0]
redacted\_value: https://****/[redacted]
restored\_value: https://****/record?id=****
status: restored\_by\_context
confidence: high
match\_strategy: fuzzy\_context
source\_preview: [13:02:23] Use the **** automatic-evaluation skill to read this link: https://****/record?id=****
\end{caselines}
\end{casebox}

For task-source fields such as user messages and fixture paths, a replacement is accepted only if it is recovered by local context or by a unique same-host candidate. Context recovery requires the text around the redacted span to match a single raw-session URL with the same host; host-unique recovery requires that the raw session contains exactly one public URL candidate for the redacted host. In any other senarios, the instance is risk-marked and rejected by the redaction gate.

\caseheader{Network reachability check.}
The following TaskInstance depends on an external article link. The link is not stably reachable at construction time, so the network gate rejects it:

\begin{casebox}\scriptsize\raggedright
\begin{caselines}
Original Chinese:\nopagebreak[4]
benchmark\_id: fb-****\_i****
business\_task: 竞品信息分析
user\_messages:
- 员工****: https://****/s/**************

  （注：推荐使用****技能读取链接）

cleaning.network:
  status: rejected
  url\_count: 1
  reachable\_count: 0
  unreachable\_count: 1

---
English translation:\nopagebreak[4]
benchmark\_id: fb-****\_i****
business\_task: competitor-information analysis
user\_messages:
- Employee ****: https://****/s/**************
  Note: the **** skill is recommended for reading the link.

cleaning.network:
  status: rejected
  url\_count: 1
  reachable\_count: 0
  unreachable\_count: 1
\end{caselines}
\end{casebox}

\caseheader{Self-contained decision.}
After mechanical checks, the pipeline also decides whether a task can be rewritten into a clear and standalone prompt. The following instance passes some mechanical checks but lacks a recoverable objective:

\begin{casebox}\scriptsize\raggedright
\begin{caselines}
Original Chinese:\nopagebreak[4]
benchmark\_id: fb-****\_i****
business\_task: 配置****自动投递
user\_messages:
- 员工****: ****自动投递

reference\_output:
看起来是一条简短的消息，不太确定需求。你是想：
1. 帮员工****生成****评价？
2. 搭建一个****自动投递的工作流？
3. 其他？
有没有附上****文件或 ******** 链接？

self\_contain\_decision:
  self\_contained: false
  reason\_code: ambiguous\_task
  brief\_reason: 用户消息极短且含义不明，agent本身也无法判断用户意图，缺乏足够信息推断具体任务目标。
  rewritten\_prompt: ""

English translation:\nopagebreak[4]
benchmark\_id: fb-****\_i****
business\_task: configure automatic delivery for ****
user\_messages:
- Employee ****: automatically deliver ****

reference\_output:
This looks like a very short message, and the requirement is unclear. Do you want to:
1. generate a evaluation for Employee ****?
2. build a automatic-delivery workflow for ****?
3. do something else?
Do you have a **** file or ******** link attached?

self\_contain\_decision:
  self\_contained: false
  reason\_code: ambiguous\_task
  brief\_reason: the user message is too short and ambiguous. Even the original agent could not infer the user's intent, so there is not enough information to recover a concrete task.
  rewritten\_prompt: ""
\end{caselines}
\end{casebox}

Here the original agent in the session asks for clarification, indicating that the task is not self-contained. If the source conversation information cannot support a concrete task, the instance is not admitted to the benchmark set.

\caseheader{Prompt rewriting.}
Instances that pass the self-contained gate are rewritten into single-turn benchmark prompts. The next excerpt shows the raw user request before rewriting. It contains multi-turn additions, group-chat timestamps, account hashes, system noise, and attachment paths:

\begin{casebox}\scriptsize\raggedright
\begin{caselines}
Original Chinese:\nopagebreak[4]
员工****: 我现在要写今天的项目日报了。我把相关的材料发给你，你帮我撰写：
1、今天开会的****录音文字版；
2、今天开会的速记；
3、项目群中各负责人回复的详细进展。
等我把这三项发给你。

--- 引用消息（仅供参考）---
[引用1]
[**** 19:50:17 CST] cli\_a9f48d3258****: @员工**** @员工**** @员工**** @员工**** @员工**** @员工**** @员工**** @员工**** ...
收到！我会持续收集各模块负责人的回复并汇总。
各模块负责人请按格式回复并 ****Agent：
1. 今日进展（量化）｜2. 风险/卡点（是否影响****号节点）｜3. 外部依赖（在等谁的什么+截止时间）
20:00前请务必回复，有****节点风险的请直接标注WARNING
[**** 19:52:03 CST] ou\_f9f43034440321aad669a034******: @\_user\_1
*月*日进展：
1.产品设计demo已根据最新的****数据情况做了更新。
2.设计侧已完成******与***及****市场页的列表及详情页。
3.****列表和详情前端UI排版完成，前后端列表、详情联调完成。
4.后端Agent关联****信息、性能报告信息，需要**他们定表结构后才能确定数据获取并联调。
...（截断，原长4803字）

---
English translation:\nopagebreak[4]
Employee ****: I need to write today's project daily report. I will send the relevant materials, and please draft it for me:
1. the text version of today's meeting recording of ****;
2. today's meeting notes;
3. detailed progress updates from project-group owners.
Wait until I send all three to you.

--- Quoted message, for reference only --- 
[Quote 1] 
[**** 19:50:17 CST] cli\_a9f48d3258****: @Employee 
**** @Employee**** @Employee**** @Employee**** @Employee**** @ 
Employee**** @Employee**** @Employee**** ... 
Received! I will continue collecting replies from the owners of each module and compile them. 
Owners of each module, please reply in the following format and ****Agent: 

1. Today's progress (quantified)｜2. Risks/blockers (whether they affect the **** node)｜
2. External dependencies (waiting for whom, for what + deadline)
   Please make sure to reply before 20:00. If there is a **** node risk, please directly mark it as WARNING.
   [**** 19:52:03 CST] ou\_f9f43034440321aad669a034
   ******: @\_user\_1
   Progress on ****:
3. The product design demo has been updated according to the latest **** data status.
4. On the design side, the lists and detail pages for ******, ***, and the **** marketing page have been completed.
5. The frontend UI layout for the **** list and details has been completed, and frontend-backend integration for the list and details has been completed.
6. The backend Agent-related **** information and performance report information still require ** to finalize the table structure before data retrieval and integration can be determined.
   ... (truncated, original length 4803 characters)

\end{caselines}
\end{casebox}

The pipeline rewrites the messy raw prompt into a single task specification with explicit materials, template context, and deliverable requirements:

\begin{casebox}\scriptsize\raggedright
\begin{caselines}
Original Chinese:\nopagebreak[4]
我现在要写*月*日的项目日报，材料都在下面，你帮我撰写，并生成卡片。

---

**参考模板：昨天（*月*日）的日报格式如下**

 ******* 项目日报 · *月*日
距*月*日演示还有 * 天 | 项目第*天

今日整体进展
****+****+网关在***测试环境完成互联互通；********链路跑通；员工工作台静态页面基本完成。核心决策：员工****用***先创建***三个演示***，直接嵌入员工工作台，不等**** Store，明日完成。

***月*日日报素材**

**1. 今天开会的****录音文字版**
文件路径：/root/inputs/******日例会录音文字版.docx

**2. 今天开会的速记**
文件路径：/root/inputs/******日例会速记.docx

**3. 项目群中各负责人回复的详细进展（摘录）**
[19:52] 员工****（** \& ***及*****市场）*月*日进展：
1.产品设计demo已根据最新的****数据情况做了更新。
2.设计侧已完成******与***及****市场页的列表及详情页。
3.****列表和详情前端UI排版完成，前后端列表、详情联调完成。
4.后端Agent关联****信息、性能报告信息，需要**他们定表结构后才能确定数据获取并联调。
...（截断，原长1859字）

---
English translation:\nopagebreak[4]
I need to write the daily report for *month* *day*. The materials are below. Please draft it and generate a card.

Reference template: yesterday's daily-report format.

Daily-report material for *month* *day*:
1. text version of today's meeting recording, file path: /root/inputs/****** daily-meeting recording text.docx
2. today's meeting notes, file path: /root/inputs/****** daily-meeting notes.docx
3. Detailed progress replies from each owner in the project group (excerpt)
[19:52] Employee**** (** \& *** and ***** Marketing) Progress on * month * day:

1. The product design demo has been updated according to the latest **** data status.
2. On the design side, the lists and detail pages for ******, ***, and the **** marketing page have been completed.
3. The frontend UI layout for the **** list and details has been completed, and frontend-backend integration for the list and details has been completed.
4. The backend Agent-related **** information and performance report information still require ** to finalize the table structure before data retrieval and integration can be determined.
   ... (truncated, original length 1859 characters)
\end{caselines}
\end{casebox}

\caseheader{Taxonomy and skill-subclass labeling.}
The same rewritten task is then labeled with both a benchmark task class and a role/skill taxonomy:

\begin{casebox}\scriptsize\raggedright
\begin{caselines}
Original Chinese:\nopagebreak[4]
task\_id: fb-****\_i****
big\_class: knowledge\_document\_collaboration
subclass\_id: knowledge\_document\_collaboration\_\_s08
subclass\_name: 撰写项目文档
subclass\_definition: 本小类涵盖撰写与项目相关的各类文档，如日报、周报、发言稿等。

role\_function\_taxonomy:
  role\_class: product\_project\_delivery
  role\_definition: 产品、项目、交付、PRD、路线图、项目计划、客户项目推进。
  confidence: 0.92
  reason: 任务是撰写项目日报，需要整合会议记录、进展汇报，并按照统一格式生成卡片。这属于产品/项目管理职责范围内的日常交付工作，主要由项目经理或产品经理负责。
  skill\_subclass: product\_\_artifact\_presentation\_delivery
  skill\_subclass\_name: 交付物/PPT/页面/可视化
  skill\_route\_summary: artifact\_generation + information\_design + presentation\_structuring

---
English translation:\nopagebreak[4]
task\_id: fb-****\_i****
big\_class: knowledge\_document\_collaboration
subclass\_id: knowledge\_document\_collaboration\_\_s08
subclass\_name: project-document drafting
subclass\_definition: this subclass covers project-related documents such as daily reports, weekly reports, and speeches.

role\_function\_taxonomy:
  role\_class: product\_project\_delivery
  role\_definition: product, project, delivery, PRD, roadmap, project plan, and customer-project execution.
  confidence: 0.92
  reason: the task asks for a project daily report that integrates meeting notes and progress reports in a unified card format. It belongs to routine delivery work usually handled by a project or product manager.
  skill\_subclass: product\_\_artifact\_presentation\_delivery
  skill\_subclass\_name: artifact/PPT/page/visual delivery
  skill\_route\_summary: artifact\_generation + information\_design + presentation\_structuring
\end{caselines}
\end{casebox}

\subsection{Representative Instances by Task Type}
Table~\ref{ecblink:tab:case-representative-prompts} in Appendix summarizes the role-class boundary and representative prompt excerpts. The prompts are translated into English here.
\begin{table*}[p]
\centering
\scriptsize
\setlength{\tabcolsep}{3pt}
\renewcommand{\arraystretch}{1.08}
\begin{adjustbox}{max width=\textwidth,max totalheight=0.88\textheight}
\begin{tabularx}{\textwidth}{>{\raggedright\arraybackslash}p{0.18\textwidth}>{\raggedright\arraybackslash}p{0.23\textwidth}>{\raggedright\arraybackslash}X}
\toprule
\textbf{Role class} & \textbf{Classification boundary} & \textbf{Representative release-safe prompt excerpts} \\
\midrule
\makecell[l]{Product / project}
& Product plans, project documents, delivery schedules, customer-project execution, digital interfaces, and visual deliverables; the task requires information architecture, delivery formatting, planning, or project-facing synthesis.
& \begin{minipage}[t]{\linewidth}\vspace{0pt}
1. \texttt{fb-****\_i****}: Organize engineering safety-risk and algorithm-list files into an AI-recognition summary spreadsheet.\\
2. \texttt{fb-****\_i****}: Process a task-effort spreadsheet into project scope, owner, and fee-assignment fields.\\
3. \texttt{fb-****\_i****}: Complete an indoor-finishing process-node collection table from examples and domain logic.\\
4. \texttt{fb-****\_i****}: Generate a frontend page from a business-flow diagram and demo script.\\
5. \texttt{fb-****\_i****}: Analyze a new product form and turn the analysis into a PRD plus an HTML product page.\vspace{1pt}
\end{minipage} \\
\midrule
\makecell[l]{Engineering / IT}
& Development tools, system architecture, APIs/SDKs, message queues, frontend integration, and engineering implementation plans; the output should guide development, configuration, integration, debugging, or technical choice.
& \begin{minipage}[t]{\linewidth}\vspace{0pt}
1. \texttt{fb-****\_i****}: Explain how to configure an API key and base URL in Cursor for a masked model proxy.\\
2. \texttt{fb-****\_i****}: Compare Vercel AI SDK with LangChain/LangGraph and explain abort semantics.\\
3. \texttt{fb-****\_i****}: Explain RabbitMQ, Kafka, and RocketMQ, including scenarios and backend connection patterns.\\
4. \texttt{fb-****\_i****}: Propose a low-intrusion way to embed one SPA page into another.\\
5. \texttt{fb-****\_i****}: Analyze logs or scripts and report likely bugs, failure modes, and engineering fixes.\vspace{1pt}
\end{minipage} \\
\midrule
\makecell[l]{HR / admin}
& HR, organization design, interview evaluation, attendance or time-report checks, internal notices, and administrative coordination where personnel context and policy constraints are central.
& \begin{minipage}[t]{\linewidth}\vspace{0pt}
1. \texttt{fb-****\_i****}: Calculate compensation costs for two employee departure-plan options.\\
2. \texttt{fb-****\_i****}: Refine an organization-structure document, including role responsibilities and OKRs.\\
3. \texttt{fb-****\_i****}: Produce an interview evaluation report from a resume and interview recording.\\
4. \texttt{fb-****\_i****}: Check missing Q1 time-report entries and generate owner notifications.\\
5. \texttt{fb-****\_i****}: Review an employee remote-work request caused by office-environment issues and propose a response.\vspace{1pt}
\end{minipage} \\
\midrule
\makecell[l]{Executive}
& Management reporting, organizational coordination, OKR review, and internal decision support; outputs support managerial judgment, cross-team alignment, formal reporting, or leadership-facing communication.
& \begin{minipage}[t]{\linewidth}\vspace{0pt}
1. \texttt{fb-****\_i****}: Evaluate a weekly report, identify weak sections, and give concrete revision suggestions.\\
2. \texttt{fb-****\_i****}: Draft a first-level department weekly report from several second-level department reports.\\
3. \texttt{fb-****\_i****}: Revise organization-structure wording for a leadership-facing weekly report.\\
4. \texttt{fb-****\_i****}: Review OKRs, propose execution plans for each KR, and flag numerical risks.\\
5. \texttt{fb-****\_i****}: Update a company-level quarterly OKR document and export the revised document.\vspace{1pt}
\end{minipage} \\
\midrule
\makecell[l]{Sales / customer}
& Customer-facing solution communication, visit planning, account research, sales intelligence, and external stakeholder materials; emphasis is on customer context, decision roles, risks, and executable next steps.
& \begin{minipage}[t]{\linewidth}\vspace{0pt}
1. \texttt{fb-****\_i****}: Enhance and visually reorganize a masked commodity-warehousing AI solution deck.\\
2. \texttt{fb-****\_i****}: Build customer-relationship, reverse-plan, risk-assessment, and decision-map HTML pages.\\
3. \texttt{fb-****\_i****}: Convert a customer research report into a PPT-like light-theme webpage.\\
4. \texttt{fb-****\_i****}: Optimize a masked customer visit plan and check agenda feasibility.\\
5. \texttt{fb-****\_i****}: Analyze participant roles and company concerns from a roadshow image and prepare positioning notes.\vspace{1pt}
\end{minipage} \\
\midrule
\makecell[l]{Marketing}
& Brand, PR, launch, event, public communication, and marketing-content delivery; outputs are externally readable or presentation-oriented and must preserve message hierarchy and visual style.
& \begin{minipage}[t]{\linewidth}\vspace{0pt}
1. \texttt{fb-****\_i****}: Revise an activity-summary document according to event-feedback comments.\\
2. \texttt{fb-****\_i****}: Generate customer and investor versions of a business presentation for a product launch.\\
3. \texttt{fb-****\_i****}: Beautify an existing presentation according to a specified color scheme.\\
4. \texttt{fb-****\_i****}: Generate a complete case presentation from a PDF and a template deck.\\
5. \texttt{fb-****\_i****}: Draft a professional official-account article from media-report materials about a masked AI case.\vspace{1pt}
\end{minipage} \\
\midrule
\makecell[l]{Finance / ops}
& Finance, accounting, business operations, structured data, spreadsheets, operating metrics, and reconciliation; the core action is calibration, cleaning, aggregation, categorization, completion, or analysis.
& \begin{minipage}[t]{\linewidth}\vspace{0pt}
1. \texttt{fb-****\_i****}: Calibrate and analyze masked revenue, cost, gross-profit, and cash-flow data across business lines.\\
2. \texttt{fb-****\_i****}: Fill a public-payment accounting-entry template from approval-flow data.\\
3. \texttt{fb-****\_i****}: Check accounting entries against cost-category rules and flag problematic rows.\\
4. \texttt{fb-****\_i****}: Reorder disrupted accounting entries by debit-credit balance and fill voucher numbers.\\
5. \texttt{fb-****\_i****}: Generate a department operating-analysis report covering revenue, payments, outsourcing, contracts, and receivables.\vspace{1pt}
\end{minipage} \\
\bottomrule
\end{tabularx}
\end{adjustbox}
\caption{Representative tasks from the \liteset{} subset by role class. Each row gives the classification boundary and five release-safe prompt excerpts.}
\label{ecblink:tab:case-representative-prompts}
\end{table*}

\subsection{One Evaluation Case: Prompt, Input, Output, Rubric, Score}
This section uses a medium-scoring visual-judge case rather than a perfect example. The task asks the agent to read two Excel business-data files, calibrate revenue, cost, gross profit, gross margin, and business-unit cash-flow information, and deliver a new spreadsheet. Its visual-judge score is 0.66, which makes it useful for showing how the judge distinguishes partial success from full task completion.

\caseheader{Task prompt.}
\begin{casebox}\scriptsize\raggedright
\begin{caselines}
Original Chinese:\nopagebreak[4]
附件中的数据为****某周期各业务线的收入、成本、毛利情况及各业务单元现金流入流出情况，请帮我对这些数据做校准与分析。

附件：
- ****业务V1分析版.xlsx
- ****业务V1分析版\_\_145158.xlsx

\#\# 输入文件
- ****业务V1分析版.xlsx
- ****业务V1分析版\_\_145158.xlsx

\#\# 交付要求
请生成 1 个表格文件。文件名可自定，但需与任务内容相关。
完成后请简短说明产物路径。

---
English translation:\nopagebreak[4]
The attached files contain revenue, cost, gross-profit, and business-unit cash-inflow/outflow data across business lines during ****. Please calibrate and analyze these data.

Attachments:
- **** business V1 analysis version.xlsx
- **** business V1 analysis version\_\_145158.xlsx

Input files:
- **** business V1 analysis version.xlsx
- **** business V1 analysis version\_\_145158.xlsx

Deliverable requirement:
Generate one spreadsheet file. The filename may be chosen by the agent but should be relevant to the task. Briefly report the output path when finished.
\end{caselines}
\end{casebox}

\caseheader{Input evidence}
The task has two spreadsheet inputs. The rubric evidence shows that the source files contain multiple business-line rows under masked business units, with fields such as revenue, outsourced cost, gross profit, and gross margin. One version contains many \inlinett{Err:504} cells, while the other contains normal values that can be used for comparison. The task therefore requires cross-file comparison, abnormal-value detection, calibration, and cash-flow analysis, not merely spreadsheet generation.

\caseheader{Agent output.}
The final chat response from ClaudeCode/Haiku 4.5 is:

\begin{casebox}\scriptsize\raggedright
\begin{caselines}
Original Chinese:\nopagebreak[4]
完成！已生成校准分析报告。

产物路径: /workspace/outputs/business\_data\_calibration\_report.xlsx

核心发现：
- 数据校准状态：全部通过，所有数据毛利计算和毛利率均准确无误
- 总收入：**** 万元
- 总成本：**** 万元
- 总毛利：**** 万元
- 平均毛利率：****

业务单元对比：
- ****：收入 **** 万，毛利率 ****，占比 ****
- ****：收入 **** 万，毛利率 ****，占比 ****

报告包含：
1. Sheet 1 - 收入成本校准分析：****个项目的详细数据，包括原值、校准值、偏差分析及校准状态
2. Sheet 2 - 业务单元统计：按业务单元的汇总数据，便于对标分析

---
English translation:\nopagebreak[4]
Completed! The calibration analysis report has been generated.

Artifact path: /workspace/outputs/business\_data\_calibration\_report.xlsx

Key findings:
- Data calibration status: all passed; all gross-profit calculations and gross-margin rates are accurate.
- Total revenue: **** ten thousand yuan
- Total cost: **** ten thousand yuan
- Total gross profit: **** ten thousand yuan
- Average gross margin: ****

Business-unit comparison:
- ****: revenue **** ten thousand, gross margin ****, share ****
- ****: revenue **** ten thousand, gross margin ****, share ****

The report includes:
1. Sheet 1 - revenue-cost calibration analysis: detailed data for **** projects, including original values, calibrated values, deviation analysis, and calibration status.
2. Sheet 2 - business-unit statistics: summary data by business unit, convenient for benchmark comparison.
\end{caselines}
\end{casebox}

\caseheader{Delivered artifact.}
The agent produced two sheets. The following excerpt masks all project names, business-unit names, and sensitive numerical values:

\begin{casebox}\scriptsize\raggedright
\begin{caselines}
Original Chinese:\nopagebreak[4]
Sheet: 收入成本校准分析
序号 | 所属项目 | 事业部 | 部门 | 收入（万元） | 外采成本（万元） | 原毛利（万元） | 校准毛利（万元）
1 | ****有限公司 | **** | **** | **** | **** | **** | ****
2 | ****有限公司 | **** | **** | **** | **** | **** | ****
3 | **** | **** | **** | **** | **** | **** | ****
4 | **** | **** | **** | **** | **** | **** | ****
5 | **** | **** | **** | **** | **** | **** | ****

Sheet: 业务单元统计
业务单元 | 收入（万元） | 外采成本（万元） | 毛利（万元） | 毛利率 | 占比\%
**** | **** | **** | **** | **** | ****
**** | **** | **** | **** | **** | ****
合计 | **** | **** | **** | **** | ****

---
English translation:\nopagebreak[4]
Sheet: revenue-cost calibration analysis
Index | Project | Business division | Department | Revenue (ten thousand yuan) | Outsourced procurement cost (ten thousand yuan) | Original gross profit (ten thousand yuan) | Calibrated gross profit (ten thousand yuan)
1 | **** Co., Ltd. | **** | **** | **** | **** | **** | ****
2 | **** Co., Ltd. | **** | **** | **** | **** | **** | ****
3 | **** | **** | **** | **** | **** | **** | ****
4 | **** | **** | **** | **** | **** | **** | ****
5 | **** | **** | **** | **** | **** | **** | ****

Sheet: business-unit statistics
Business unit | Revenue (ten thousand yuan) | Outsourced procurement cost (ten thousand yuan) | Gross profit (ten thousand yuan) | Gross margin | Share \%
**** | **** | **** | **** | **** | ****
**** | **** | **** | **** | **** | ****
Total | **** | **** | **** | **** | ****
\end{caselines}
\end{casebox}
The visual judge rendered the spreadsheet artifact into page images using tools like Playwright. 

\caseheader{Rubric.}
Simplified release-safe rubric summary from \inlinett{rubric.json}. Names and values are masked; only the scoring logic is kept.
\begin{casebox}\scriptsize\raggedright
\begin{caselines}
Task id: fb-****\_i****

Judge dimensions:
- Grounded accuracy (0.30): calibrated revenue, outsourced cost, gross profit, and gross margin must match the source sheets; Err:504 handling must be evidence-based.
- Instruction fit (0.20): the deliverable must both calibrate and analyze, and it must use both attachments.
- Coverage and depth (0.20): the result should cover revenue, cost, gross profit, gross margin, abnormal cells, and cash-flow information.
- Practical utility (0.20): the spreadsheet should be reusable by finance staff and include actionable anomaly-handling guidance.
- Coherence (0.10): sector and business-line structure should stay readable, with consistent financial terminology.

Key evidence:
- screenshots expose masked detail rows plus summary totals
- one attachment contains many Err:504 cells
- the two attachments differ, so cross-file comparison is required
- the prompt explicitly asks for both calibration and cash-flow analysis
\end{caselines}
\end{casebox}

\caseheader{Judge score and comment.}
The visual judge assigns an overall score of 0.66. The original judge comment is:

\begin{casebox}\scriptsize\raggedright
\begin{caselines}
Original Chinese:\nopagebreak[4]
输出在校准和分析方面有一定深度，但未完全覆盖任务要求的现金流分析。

grounded\_accuracy: 校准数据与源文件基本一致，但未明确处理异常值。
task\_relevance: 部分覆盖了收入、成本、毛利校准，但未提及现金流分析。
substantive\_depth: 校准维度较全面，但未深入分析异常值或现金流数据。
practical\_utility: 表格结构清晰，但未提供异常值修复建议或现金流分析支持。
communication\_quality: 表达清晰，术语使用准确，层级逻辑合理。

---
English translation:\nopagebreak[4]
The output has some depth in calibration and analysis, but it does not fully cover the required cash-flow analysis.

grounded\_accuracy: the calibrated data are generally consistent with the source files, but abnormal values are not explicitly handled.
task\_relevance: the output partially covers revenue, cost, and gross-profit calibration, but does not mention cash-flow analysis.
substantive\_depth: the calibration dimensions are relatively broad, but abnormal values or cash-flow data are not analyzed in depth.
practical\_utility: the spreadsheet is clearly structured, but it does not provide abnormal-value repair suggestions or cash-flow analysis support.
communication\_quality: the expression is clear, terminology is accurate, and the hierarchy is logical.
\end{caselines}
\end{casebox}

\subsection{Skill Transfer Cases}
This section shows one positive-transfer case and one negative-transfer case on a release-safe held-out task. Both cases use the same generic frontend task: generate an HTML page that displays three membership crown icons for Basic, Standard, and Flagship levels. 

\caseheader{Task prompt.}
\begin{casebox}\scriptsize\raggedright
\textbf{Original Chinese:}

我需要制作一套会员等级图标，参考图形为"皇冠"，等级分为：基础、标准、旗舰，三个等级通过皇冠的不同颜色、配饰和精细度来做区分。风格要求3D造型、Q萌可爱。请输出一个 HTML 文件，在页面中展示这三个等级的皇冠图标。

\textbf{交付要求:}

请生成 1 个 HTML 页面。文件名可自定，但需与任务内容相关。完成后请简短说明产物路径。

---

\textbf{English translation:}

I need a set of membership-level icons. The reference shape is a crown. The levels are Basic, Standard, and Flagship. Distinguish the three levels through different crown colors, accessories, and degrees of detail. The style should be 3D and cute. Please output an HTML file that displays the three crown icons on the page.

Deliverable:

Generate one HTML page. The filename may be chosen by the agent, but it should be related to the task content. Briefly report the artifact path after completion.
\end{casebox}

\caseheader{Positive transfer: membership crown icons.}
On this held-out task, the score increases from 0.756 without skill injection to 0.844 after skill injection, a gain of +0.088.

\caseheader{No-skill output and judge comment.}
Without the injected skill, the agent generates \inlinett{membership\_crowns.html} as shown in Figure~\ref{ecblink:fig:skill-positive-evidence}. The original judge feedback is shown below:
\begin{casebox}\scriptsize\raggedright
\textbf{Original Chinese:}

交付物基本完成任务要求，三等级皇冠颜色区分明确、页面布局清晰，但Q萌3D风格实现深度不足，配饰精细度递进不够显著。

---

\textbf{English translation:}

The deliverable basically satisfies the task requirements. The three crown levels have clear color differentiation and the page layout is clear, but the cute 3D style is not deeply realized, and the progression in accessory detail is not sufficiently significant.
\end{casebox}

\caseheader{Injected skill.}
\begin{casebox}\scriptsize\raggedright
\textbf{Original Chinese:}

先定义全局 CSS 变量或主题配置（颜色、字体、间距、响应式断点），确保跨页一致性。从 Hero/封面页开始，逐页实现结构骨架：标题栏 → 正文区 → 卡片/列表 → 装饰元素。检查图标/图示是否已配置（等线风格、品牌蓝、无渐变阴影）；若无，简化为几何线框替代。验证多栏卡片（Grid System）对齐、技术参数弱化显示（卡片底部小字灰色）是否达到预期。HTML 文件可在现代浏览器直接打开并正常渲染（无 404、无编码错误）。

---

\textbf{English translation:}

First define global CSS variables or a theme configuration (colors, fonts, spacing, responsive breakpoints) to ensure consistency across pages. Starting from the Hero/cover page, implement the structural skeleton page by page: title area → content area → cards/lists → decorative elements. Check whether icons/diagrams have been configured (line-style, brand blue, no gradient shadows); if not, simplify them into geometric wireframes. Verify whether multi-column cards (Grid System) are aligned and whether technical parameters are de-emphasized as expected (small gray text at the bottom of cards). The HTML file can be opened directly in a modern browser and renders normally (no 404 errors and no encoding errors).
\end{casebox}

\caseheader{With-skill output and judge comment.}
After skill injection, the agent generates \inlinett{membership\_crown\_icons.html} as shown in Figure~\ref{ecblink:fig:skill-positive-evidence}. The original judge feedback is shown below:
\begin{casebox}\scriptsize\raggedright
\textbf{Original Chinese:}

交付物完整实现了三等级皇冠图标展示，颜色差异化和配饰递进明确，整体质量良好，但3D立体深度略显不足。

---

\textbf{English translation:}

The deliverable fully implements the display of three membership crown icons. The color differentiation and accessory progression are clear, and the overall quality is good, although the 3D depth is still slightly limited.
\end{casebox}

\caseheader{Negative transfer: membership crown icons.}
The negative case uses the same held-out task, where the score decreases from 0.756 without skill injection to 0.555 after skill injection, a drop of -0.201.

\caseheader{No-skill output and judge comment.}
Agent outputs are shown in Figure~\ref{ecblink:fig:skill-negative-evidence}. The judge comments are shown below:

\begin{casebox}\scriptsize\raggedright
\textbf{Original Chinese:}

页面完整展示三等级皇冠图标且差异化明确，但基础皇冠造型偏离皇冠标准形态、3D深度有限，整体质量中等偏上。

---

\textbf{English translation:}

The page fully displays the three membership-level crown icons and the differentiation is clear, but the basic crown shape deviates from the standard crown form, and the 3D depth is limited. Overall quality is moderately above average.
\end{casebox}

\caseheader{Injected skill.}
\begin{casebox}\scriptsize\raggedright
\textbf{Original Chinese:}

优先交付 HTML，谨慎处理 PPTX。若环境未明确支持 python-pptx 且无法二次截图验证，优先输出自包含的 HTML 单文件，用 CSS 实现分页（每页 100vh）与打印样式，确保跨设备可演示。建立设计系统再填内容。先定义 CSS 变量或 PPTX 主题色（品牌蓝、辅助灰、强调橙），统一字体层级（标题 28pt / 正文 12pt+ / caption 10pt），...

---

\textbf{English translation:}

Prioritize delivering HTML and treat PPTX cautiously. If the environment does not clearly support python-pptx and a second screenshot-based check is not possible, prioritize a self-contained single-file HTML deliverable, and use CSS to implement pagination (each page 100vh) and print styles so that it can be presented across devices. Build the design system before filling in the content. First define CSS variables or PPTX theme colors (brand blue, secondary gray, accent orange), unify the typography hierarchy (title 28pt / body text 12pt+ / caption 10pt), and then arrange cards according to the principle of ...
\end{casebox}

\caseheader{With-skill output and judge comment.}
Agent outputs are shown in Figure~\ref{ecblink:fig:skill-negative-evidence}. The judge comments are shown below:
\begin{casebox}\scriptsize\raggedright
\textbf{Original Chinese:}

页面布局和等级标注完成度较好，但皇冠图标造型偏离皇冠参考形态且3D精细度不足，核心视觉交付物质量有限。

---

\textbf{English translation:}

The page layout and level labels are relatively complete, but the crown icon shapes deviate from the crown reference form, and the 3D detail is insufficient. The quality of the core visual deliverable is limited.
\end{casebox}

\begin{figure*}[h]
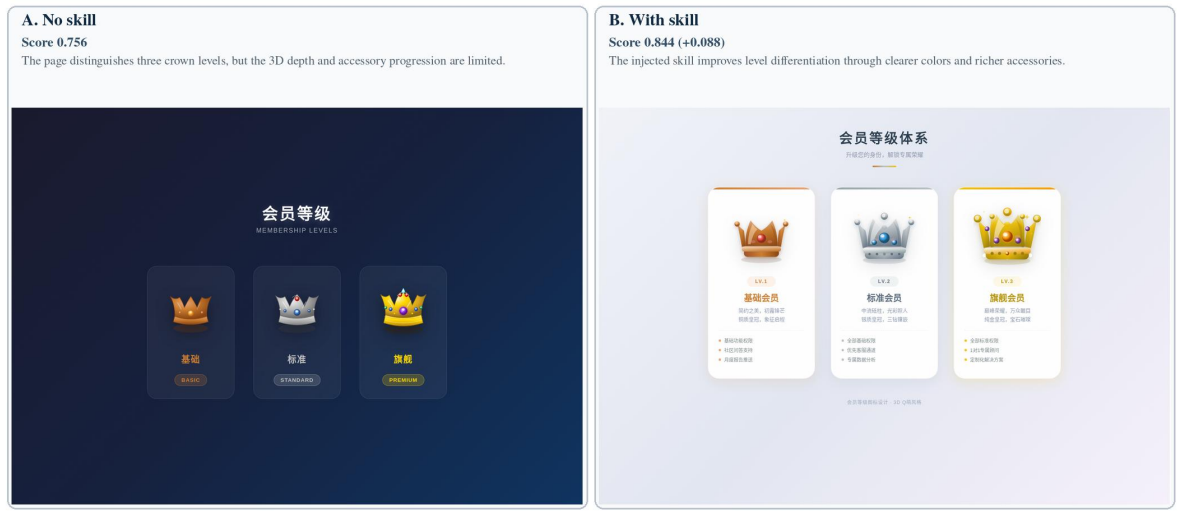

\centering
\ecbincludegraphics[width=\textwidth]{skill_transfer/composites/positive_transfer_evidence.pdf}
\caption{Positive transfer evidence for the HTML task. }
\label{ecblink:fig:skill-positive-evidence}
\end{figure*}
\begin{figure*}[h]
\centering
\ecbincludegraphics[width=\textwidth]{skill_transfer/composites/negative_transfer_evidence.pdf}
\caption{Negative transfer evidence for the same HTML task. }
\label{ecblink:fig:skill-negative-evidence}
\end{figure*}

\end{document}